%% file: emnlp_2025_camera_ready.tex
\pdfoutput=1

\documentclass[11pt]{article}

\usepackage[final]{acl}

\usepackage{times}
\usepackage{latexsym}

\usepackage[T1]{fontenc}

\usepackage[utf8]{inputenc}

\usepackage{microtype}

\usepackage{inconsolata}

\usepackage{graphicx}

\usepackage[utf8]{inputenc} 
\usepackage[T1]{fontenc}    
\usepackage{hyperref}       
\usepackage{url}            
\usepackage{booktabs}       
\usepackage{amsfonts}       
\usepackage{amssymb}
\usepackage{nicefrac}       
\usepackage{microtype}      
\usepackage[dvipsnames]{xcolor}         
\usepackage{multirow}
\usepackage[breakable]{tcolorbox}
\usepackage{colortbl}
\usepackage{natbib}
\usepackage{listings}

\input{math_commands.tex}

\usepackage{tabularx}
\usepackage{subcaption}
\usepackage{graphicx}
\usepackage{enumitem}
\setlist{nosep}
\usepackage{soul}
\usepackage{pifont}
\usepackage{makecell}
\usepackage{longtable}
\usepackage{xspace}
\usepackage{colortbl}
\usepackage{float}
\newfloat{listing}{tbp}{lst}[section]
\floatname{listing}{Listing}

\definecolor{background}{rgb}{0.95, 0.95, 0.92}
\lstdefinestyle{mystyle}{  
    backgroundcolor=\color{background},
    basicstyle=\ttfamily\scriptsize,
    breakatwhitespace=false,         
    breaklines=true,                 
    captionpos=b,                    
    keepspaces=true,                 
    numbers=left,                    
    numbersep=5.5pt,                  
    showspaces=false,                
    showstringspaces=false,
    showtabs=false,                  
    tabsize=2,
    xleftmargin=.05\textwidth, 
    xrightmargin=.05\textwidth
}
\lstdefinelanguage{json}{
    backgroundcolor=\color{background},
    basicstyle=\scriptsize\ttfamily,
    stepnumber=1,
    numbers=left,
    numbersep=5.5pt,
    showstringspaces=false,
    breaklines=true,
    xleftmargin=.05\textwidth, 
    xrightmargin=.05\textwidth
}

\newcommand{\cmark}{\ding{51}}
\newcommand{\xmark}{\ding{55}}

\newcommand{\yfl}[1]{{\color{SeaGreen}#1}}
\newcommand{\myred}[1]{{\color{OrangeRed}#1}}
\newcommand{\mygreen}[1]{{\color{ForestGreen}#1}}

\newcommand{\pipeline}{AutoSDT\xspace}

\newcommand{\dataset}{AutoSDT-5K\xspace}

\title{AutoSDT: Scaling Data-Driven Discovery Tasks Toward Open Co-Scientists}

\author{Yifei Li$^{1,\dagger}$\thanks{Equal Contribution. See contribution statement for details.}, \textbf{Hanane Nour Moussa$^{1,\dagger}$\footnotemark[\value{footnote}],} \textbf{Ziru Chen$^{1,\dagger}$,} \textbf{Shijie Chen$^{1,\dagger}$,} \textbf{Botao Yu$^{1,\dagger}$,} \\
\textbf{Mingyi Xue$^{6,\diamond}$,}
\textbf{Benjamin Burns$^{1,\dagger}$,} \textbf{Tzu-Yao Chiu$^{3,\dagger}$,} \textbf{Vishal Dey$^{1,\dagger}$,} \textbf{Zitong Lu$^{3,\dagger}$,} \\
\textbf{Chen Wei$^{5,\diamond}$,} \textbf{Qianheng Zhang$^{5,\diamond}$,}
\textbf{Tianyu Zhang$^{3,\dagger}$,} \textbf{Song Gao$^{5,\diamond}$,} \textbf{Xuhui Huang$^{6,\diamond}$,} \\
\textbf{Xia Ning$^{1,2,4,\dagger}$,} \textbf{Nesreen K. Ahmed$^\circ$,} \textbf{Ali Payani$^\circ$,} \textbf{Huan Sun$^{1,\dagger}$}\\
$^1$Department of Computer Science and Engineering \quad $^2$College of Pharmacy \\
$^3$Department of Psychology \quad $^4$Department of Biomedical Informatics \\
$^5$Department of Geography \quad $^6$Department of Chemistry \\
$\circ$Cisco Research \quad
$\diamond$University of Wisconsin–Madison \\
$\dagger$The Ohio State University
}
\begin{document}

\maketitle

\begin{abstract}
Despite long-standing efforts in accelerating scientific discovery with AI, building AI co-scientists remains challenging due to limited high-quality data for training and evaluation. To tackle this data scarcity issue, we present \pipeline, an automatic pipeline that collects high-quality coding tasks in real-world data-driven discovery workflows. \pipeline leverages the coding capabilities and parametric knowledge of LLMs to search for diverse sources, select ecologically valid tasks, and synthesize accurate task instructions and code solutions. Using our pipeline, we construct \dataset, a dataset of 5,404 coding tasks for data-driven discovery that covers four scientific disciplines and 756 unique Python packages. \textit{To the best of our knowledge, AutoSDT-5K is the only automatically collected and the largest open dataset for data-driven scientific discovery}. Expert feedback on a subset of 256 tasks shows the effectiveness of \pipeline: 93\% of the collected tasks are ecologically valid, and 92.2\% of the synthesized programs are functionally correct. Trained on \dataset, the Qwen2.5-Coder-Instruct LLM series, dubbed AutoSDT-Coder, show substantial improvement on two challenging data-driven discovery benchmarks, ScienceAgentBench and DiscoveryBench. Most notably, AutoSDT-Coder-32B reaches the same level of performance as GPT-4o on ScienceAgentBench with a success rate of 7.8\%, doubling the performance of its base model. On DiscoveryBench, it lifts the hypothesis matching score to 8.1, bringing a 17.4\% relative improvement and closing the gap between open-weight models and GPT-4o.\footnote{Our code and data are publicly available at \href{https://osu-nlp-group.github.io/AutoSDT/}{https://osu-nlp-group.github.io/AutoSDT/.}}
\end{abstract}

\section{Introduction}
\label{sec:intro}

Accelerating scientific research and development has long been an aspirational goal in AI research \citep{langley1987scientific, doi:10.1126/science.1172781}.
Since the 1960s, there have been continuous efforts in developing AI methods for scientific discovery, e.g., by constructing massive knowledge bases \citep{BUCHANAN19785}, designing expert rules and heuristics \citep{10.5555/1623156.1623181}, and learning representations from large-scale data \citep{Jumper2021}.
Nonetheless, they are tailored for very specific tasks with constrained solution spaces.
The realization of an AI assistant for open-ended scientific discovery still seems distant.

\input{figures/highlight}

Recently, large language models (LLMs) bring new light to fulfill this ambition and
have piqued significant interest in building ``AI co-scientists'' \citep{coscientist, gottweis2025aicoscientist} that assist in scientific discovery.
Specifically, due to their digital nature, most AI co-scientist agents focus on data-driven discovery workflows \citep{hey2009fourthparadigm, majumder2024position}, including scientific computation and analysis \citep{tian2024scicode, chen2025scienceagentbench}, symbolic regression \citep{shojaee2025llmsr, shojaee2025llmsrbenchnewbenchmarkscientific}, and hypothesis generation \citep{majumder2025discoverybench, mitchener2025bixbenchcomprehensivebenchmarkllmbased}.
While some agents have shown promising results \citep{gottweis2025aicoscientist, lu2024aiscientistfullyautomated}, they mostly rely on proprietary LLMs, which impedes their adoption in subjects that require transparency and data privacy, such as social science \citep{Ollion2024dangerllm} and medicine \citep{Zhang2024closegap}.
Thus, there is a strong need for AI co-scientists powered by open-weight LLMs. 

To build AI co-scientists, one critical bottleneck is the absence of large-scale, high-quality data for training and evaluation.
Commonly formulated as code generation problems \citep{chen2025scienceagentbench, majumder2025discoverybench, mitchener2025bixbenchcomprehensivebenchmarkllmbased}, data-driven discovery tasks require AI agents to derive scientific insights by processing, analyzing, and visualizing data.
On one hand, automatically mining data-driven discovery tasks is particularly challenging.
Unlike software engineering tasks \citep{jimenez2024swebench}, which can often be extracted from code changes in pull requests, \textit{data-driven discovery tasks require complete, file-level code that operates on real-world scientific datasets and solves domain-specific problems}, which cannot be directly crawled from existing code repositories.
On the other hand, manual task annotation is quite time-consuming. 
It takes trained graduate students at least 2.5--3 hours to annotate one task \citep{chen2025scienceagentbench, majumder2025discoverybench}, not to mention extra time for paper searching and task validation.

In this paper, we present \pipeline, a pipeline for \underline{auto}matically \underline{s}caling \underline{d}ata-driven discovery \underline{t}asks to tackle the data scarcity issue from three aspects.
\textbf{(1) Source Diversity}: Our pipeline overcomes the lack of source diversity in manually annotated datasets \citep{chen2025scienceagentbench, majumder2025discoverybench} by using LLM-based query augmentation to systematically search for code repositories that contain data-driven discovery tasks.
\textbf{(2) Task Ecological Validity}: We exploit LLMs' parametric knowledge to locate programs that resemble real-world data-driven discovery tasks and generate scientifically accurate task instructions.
\textbf{(3) Code Quality}: Each selected program goes through multiple rounds of adaptation and validation by an LLM to ensure standalone executability and functional equivalence to the original code.

We use our pipeline to create \dataset, a dataset of 5,404 data-driven discovery tasks, which costs only 0.55 USD per task on average.
To our best knowledge, \dataset is so far the largest coding dataset for data-driven discovery, covering 756 unique Python packages of computational tools in four disciplines: Bioinformatics, Computational Chemistry, Geographical Information Science, and Psychology and Cognitive Neuroscience.
We also engage 9 subject matter experts from these disciplines, including Ph.D. students and professors, to examine a subset of 256 tasks.
The experts report that \textbf{93\%} of the tasks are scientifically authentic and represent parts of their data-driven discovery workflows, and \textbf{92.2\%} of the generated programs are deemed correct solutions to the tasks, validating the high quality of \dataset.

\input{figures/AutoDCT}

Through comprehensive experiments, we further demonstrate the utility of \dataset.
We fine-tune Qwen2.5-Coder-Instruct \citep{hui2024qwen25codertechnicalreport} on \dataset and obtain AutoSDT-Coder, a series of LLMs with improved coding capabilities for data-driven discovery. 
We evaluate AutoSDT-Coder on two challenging data-driven discovery benchmarks, ScienceAgentBench \citep{chen2025scienceagentbench} and DiscoveryBench \citep{majumder2025discoverybench}.
As shown in Figure \ref{fig:highlight}, AutoSDT-Coder-32B reaches the same level of performance as GPT-4o (2024-05-13) with a 7.8\% success rate (\textbf{SR}) on ScienceAgentBench, double the performance of the base model  (3.9\% \textbf{SR}).
On DiscoveryBench, AutoSDT-Coder-32B also brings a 17.4\% relative improvement over its base LLM, lifting the hypothesis matching score from 6.9 to 8.1.
These results illustrate how \pipeline can propel the advancement toward open AI co-scientists by automatically scaling high-quality data-driven discovery tasks.
\vspace{-0.5pt}

\section{\pipeline}
\label{sec:pipeline}


Since manual annotation requires extensive labor and high expertise, existing data-driven discovery datasets \citep{chen2025scienceagentbench, majumder2025discoverybench} contain only a few hundred tasks for evaluation only.
To enable LLM training with a reasonable amount of data, we propose \pipeline, a fully automatic pipeline for collecting data-driven discovery tasks at scale.
As shown in Figure \ref{fig:AutoDCT}, given a few high-level keywords, \pipeline-Search first searches for related code repositories with keyword expansion (Section \ref{sec:crawl}). 
After that, \pipeline-Select identifies source code files that correspond to data-driven discovery tasks with multi-step filtering
and extracts dependencies for their execution environments (Section \ref{sec:select}). 
Lastly, \pipeline-Adapt modifies the selected source code files into independently executable programs and generates task instructions accordingly (Section \ref{sec:adapt}).





\subsection{\pipeline-Search}
\label{sec:crawl}


\pipeline starts with searching for code repositories containing programs for data-driven discovery tasks.
This process is initiated with user-provided keywords describing the topic of interest, such as ``bioinformatics,'' which is the only human effort needed in \pipeline.
These keywords are subsequently expanded into a comprehensive set of related search queries (e.g., ``genomics'' and ``molecular data'') by an LLM\footnote{LLM is GPT-4o (2024-11-20) unless otherwise indicated.}, which significantly improves the coverage of the search results. For example, in our preliminary attempts, we can only find 332 repositories using the keyword ``neuroscience'' alone. However, after expanding the keyword into a list containing ``neuroimaging,'' ``neuroplasticity,'' and ``neuroinformatics,'' \pipeline-Search can double the number of its identified repositories in this discipline to 693.

 


We consider two popular code hosting platforms among researchers, GitHub and PapersWithCode, and use their search APIs to collect a list of repositories.
After that, we use an LLM to judge whether each repository indeed hosts code related to a research paper in the targeted discipline according to the \texttt{README.md} file (prompt in Appendix Table \ref{tab:crawl-prompt}).
\textit{We then eliminate duplicate repositories and ensure that there is no overlap with the repositories utilized in existing benchmarks \citep{chen2025scienceagentbench, majumder2025discoverybench} which we use for evaluation}. 
This process yields a large collection of code repositories related to the disciplines of interest, which are further processed by subsequent steps in the pipeline.
\subsection{\pipeline-Select}
\label{sec:select}
\pipeline processes the crawled repositories with three sub-steps
to identify source code files of data-driven discovery tasks and prepare their execution environments (workspaces).

\noindent \textbf{Crawling Python Files.} \pipeline clones the identified repositories and first extracts all Python files. Then, it applies rule-based filtering to remove files exceeding 1,000 lines and directories that are unlikely to contain substantive data-driven discovery tasks (e.g., ``config'' and ``tests'').


\noindent \textbf{Data-driven Scientific Code Filtering.} 
We leverage the LLM's parametric knowledge to determine the filtered source code files' relevance to data-driven discovery.
Specifically, the LLM assesses if each file meets three criteria: (1) its functionality is related to data-driven scientific workflows, such as model training, computational analysis, and data visualization; (2) it utilizes one or more datasets as program inputs; and (3) it generates scientific outputs such as numerical results, processed datasets, or visualizations. The prompt used for this step is in Appendix Table \ref{tab:scitaskverify}. 

\noindent \textbf{Dependency Extraction and Workspace Preparation.} The third sub-step focuses on extracting all necessary dependencies for executing the programs, including datasets, pre-trained models, and auxiliary utility code. 
This process leverages the LLM's code understanding capabilities to automatically identify required dependencies. 
The LLM analyzes both file content and repository structure to recognize all dependencies within the repository and returns a list of their paths. 
We provide the prompt used for dependency extraction in Appendix Table \ref{tab:locate}. 
This step allows us to prepare compact workspaces by only storing necessary files. 
As a result, the average size of workspaces is only 40.42 MB as opposed to 264.98 MB of repositories.


\subsection{\pipeline-Adapt}
\label{sec:adapt}

Finally, \pipeline-Adapt creates <task instruction, code solution> pairs by adapting the identified code snippets into independently executable programs and generating corresponding task instructions.

\noindent \textbf{Program Adaptation.} Program files taken directly from code repositories often fail to run locally for many reasons such as dependency issues, missing configurations, or other implementation bugs. Therefore, we introduce a code adaptation process to convert raw code files from the repository into standalone and executable code in three sub-steps: 
(1) We first prompt Claude-3.7-Sonnet \citep{anthropic2025claude3.7} to generate an initial adaptation of the source code.\footnote{In most stages of \pipeline, we use GPT-4o (2024-11-20) for its general capabilities and lower cost but adopt Claude-3.7-Sonnet for its high coding performance.}
Given the source code and workspace structure, the LLM is supposed to make modifications to import statements, input/output routines, and hard-coded paths without changing the program's core functionality (Appendix Table \ref{tab:adapt}).
(2) After obtaining the adapted program, we use \texttt{pipreqs}\footnote{https://github.com/bndr/pipreqs} to extract Python dependencies and prepare the conda environments for execution.
(3) The adapted program is then executed in the configured conda environments for self-debugging \citep{chen2024self_debug}.
We repeat the program generation and execution loop for at most three iterations and discard those still having execution errors after the loop finishes.

\noindent \textbf{Task Instruction Generation.}
Given an adapted program, we prompt an LLM to back-translate it into a clear task instruction (Appendix Table \ref{tab:instructiongen}) that explicitly includes the task goal, required input data and/or model files, and expected output files (examples in Appendix Table \ref{tab:examples}). 
According to expert feedback (Section \ref{sec:eval}), most of the generated instructions are correctly expressed in subject-specific scientific language. 
With programs adapted and instructions generated, we obtain <task instruction, code solution> pairs that represent real-world tasks in data-driven scientific discovery. An example task is provided in Appendix \ref{app:examples}. 
\input{tables/dataset_statistics}


\section{\dataset}
\label{sec:dataset}


\subsection{Statistics}
\label{sec:stats}

\input{figures/subtasks_and_packages}

We apply \pipeline to collect data-driven discovery tasks in four disciplines: Bioinformatics, Chemistry, Geographical Information Science, and Psychology and Cognitive Neuroscience. 
\pipeline successfully retrieves 2,993 research-related repositories, selects files from 1,325 of them for further processing, and synthesizes 5,404 coding tasks to compose \dataset (Table \ref{tab:dataset_statistics}), a large-scale dataset for data-driven discovery.

\noindent\textbf{Coverage.} \dataset covers diverse types of data-driven discovery tasks, where each task comprises multiple subtasks representing specific workflow components (e.g., data transformation, model training, visualization), 
as illustrated in Figure \ref{fig:subtasks}.  
In terms of required tools, in addition to common data analysis packages such as \texttt{sklearn} and \texttt{scipy}, \dataset also includes a wide range of domain-specific packages, such as  \texttt{ase} for atomic simulations,  \texttt{nibabel} for reading neuroimaging files, and \texttt{geopandas} for handling geospatial data. Figure \ref{fig:packages} shows representative general and domain-specific packages and their occurrence in \dataset.

\noindent\textbf{Cost.} In processing all 2,993 repositories and generating 5.4K tasks, our pipeline incurs a total API cost of 2,955 USD, with a cost per task as low as 0.55 USD. 
The detailed breakdown of the cost is given in Appendix \ref{app:cost}.
For reference, the human effort required to annotate a similar task is 2.5-3 hrs per \citet{chen2025scienceagentbench}, which translates to at least 20 USD per task using minimum annotator rates\footnote {https://researcher-help.prolific.com/en/article/9cd998}. 
\input{tables/analysis_per_difficulty_level}

\subsection{Expert Evaluation}
\label{sec:eval}


To confirm the quality of the tasks in \dataset, we involve nine subject matter experts to conduct a rigorous evaluation: 3 in bioinformatics and computational chemistry, 3 in geographic information science, and 3 in psychology and cognitive neuroscience. 
We randomly sample 256 tasks from \dataset (96 in bioinformatics/computational chemistry, 75 in geographic information science, and 85 in psychology \& cognitive neuroscience). For each task, we prepare a folder containing the task instruction, the link to the original file on GitHub, the program solution, and the program dependencies. The questionnaire used for task evaluation is given in Appendix Table \ref{box:questionnaire}.


\noindent \textbf{Task Instruction Validity.} As judged by domain experts, \textbf{93\% }of the instructions in \dataset describe meaningful tasks that scientists would encounter in their day-to-day research activities. 
In addition, \textbf{91.4\%} of the task instructions are correctly expressed in the domain scientific language, adding to the ecological validity of the dataset.
\textbf{73.4\%} of the instructions
are also clear and contain all the required information to solve the task, namely the task goal, input data, and expected output, showing the effectiveness of our instruction generation pipeline. For the 26.7\% of task instructions in which the clarity is lacking, the expert feedback suggests that it is mainly due to the lack of detailed guidance about the methods to be used to solve the task. For example, a task in geographic information science specifies the goal of calculating the Rossby radius of deformation and provide the latitude and longitude of the geographic coordinate, however, in order to perform this calculation other parameters are needed such as buoyancy. 

This issue may stem from limited context provided to the LLM for some code files found on GitHub that are often not well documented. A meaningful direction of future exploration would be the incorporation of additional information from the code repository or related publication in the instruction generation stage. However, careful consideration must be given to selecting the context that improves the model's understanding of the task background without overloading it with irrelevant information. We further discuss limitations and future work in the \hyperref[sec:limitations]{Limitations} section.
\input{tables/main_results_sab_v2_expanded_nocbs.tex}

\noindent\textbf{Code Solution Correctness.} The expert evaluation further confirms the effectiveness of \pipeline in code adaptation.
\pipeline is able to successfully modify the code for standalone executability without altering its original functionality \textbf{84.4\%} of the time. Based on expert feedback, many of the cases where the codes are not completely equivalent are due to the adapted code locally implementing missing dependencies. However, the correctness of the adapted code is still high, with \textbf{92.2\%} of programs deemed correct solutions to their task instructions.

\noindent \textbf{Task Difficulty.} In judging the task difficulty, the experts were instructed to estimate how much time it would take them to write a code solution to the task instruction. Similar to \citet{yang2025swesmithscalingdatasoftware}, tasks that have a completion time of \(\leq\)15 min are deemed \textit{Easy}, those requiring 15 min -- 1 hr are \textit{Medium}, and tasks requiring 1+ hrs are \textit{Hard}. As illustrated in Table \ref{tab:analysis-difficulty-levels}, the expert ratings show varied difficulty levels, with more than \textbf{75\% }of the tasks falling in the \textit{Medium} to \textit{Hard} range. On average, the \textit{Hard} tasks contain considerably more lines of code and more subtasks.

\yfl{\paragraph{Comparison with related datasets.} }

\noindent Overall, \dataset is a large-scale and high-quality dataset for data-driven discovery tasks, making it a valuable resource for developing future co-scientist agents. Compared to related datasets \cite{chen2025scienceagentbench, majumder2025discoverybench, gu2024bladebenchmarkinglanguagemodel, mitchener2025bixbenchcomprehensivebenchmarkllmbased}, \dataset covers a considerably larger set of tasks balanced across multiple disciplines and is the largest open dataset for data driven scientific discovery and the only automatically collected one to the best of our knowledge. \textit{\dataset is also the only dataset that is large enough to be used for training purposes, whereas other datasets are used for evaluation only.} \dataset is adapted from naturally-occurring scientist-authored code, ensuring that the tasks represent genuine scientific workflows (see Appendix \ref{app:comparison}).


\section{Experiments}
\label{sec:experiments}
\subsection{Experimental Setup}

\noindent \textbf{Datasets.}
We conduct experiments to show the effectiveness of training on \dataset using two data-driven discovery benchmarks: ScienceAgentBench \citep{chen2025scienceagentbench}
and DiscoveryBench \citep{majumder2025discoverybench}.
In ScienceAgentBench, given a task instruction and dataset information, a method is supposed to generate a complete Python program to solve the task in an end-to-end manner. This requires the generated program to correctly process the input data, implement correct functionality to model, analyze, or visualize the data, and finally save the results into the correct output path.
In DiscoveryBench, 
given the description of input data schema and a scientific query, a method is supposed to first generate Python code to analyze the data based on the query and then generate scientific hypotheses.

\noindent \textbf{Models.}
We choose Qwen2.5-Coder-Instruct series \citep{hui2024qwen2} as our base models for supervised fine-tuning, due to their superior performance on existing coding benchmarks \citep{yang2025swesmithscalingdatasoftware, jain2025r2egymproceduralenvironmentshybrid, xie2025repostscalablerepositorylevelcoding}.
On ScienceAgentBench, we evaluate: (1) five open-weight LLMs: Llama-3.1-Instruct-70B, 405B \citep{grattafiori2024llama}, and Qwen2.5-Coder-Instruct-7B, 14B, and 32B \citep{hui2024qwen25codertechnicalreport}.
On DiscoveryBench, we compare performance with GPT-4o (2024-11-20) and Qwen2.5-Coder-Instruct. 
We re-implemented some inference steps to decouple code generation from hypothesis generation, partially leading to different results reproduced in our paper. For all the inferences, we use a temperature of $0.2$ and top\_p of $0.95$, and perform 0-shot direct prompting. More details of training and inference settings can be found in Appendix \ref{app:training-details}.

\noindent \textbf{Evaluation metrics.}
For ScienceAgentBench, we report (1) Success Rate (\textbf{SR}): a binary metric that examines whether a program output meets the human-annotated success criteria for each task goal and (2) Valid Execution Rate (\textbf{VER}): a binary metric which checks if the program can execute without errors and save its output to the correct location.
In DiscoveryBench, we evaluate the generated hypotheses against gold hypotheses using Hypothesis Matching Score (\textbf{HMS}), which breaks them down into sub-hypotheses with GPT-4o (2024-11-20)\footnote{The original evaluator LLM, gpt-4-preview-0125, is no longer available since 05/01/2025.} to calculate semantic matches.
For both datasets, we sample 3 responses and report the average score (with standard deviation) for all metrics. We put more implementation details in Appendix \ref{app:training-details}.



\subsection{Main Results}


\textbf{Training on \dataset effectively improves the performance on data-driven discovery tasks.} Table \ref{tab:direct_prompting_results} demonstrates that models trained on \dataset achieve improved SR and VER on ScienceAgentBench. 
Specifically, we improve Qwen2.5-Coder-32B by 3.9\% SR and 7.6\% VER. Furthermore, we notice that the performance gains increase with model size; although the 7B model does not show SR improvements, the performance gains become more evident with the 14B and 
 the 32B model. The performance drop for the 7B model is probably due to limited model capacity \citep{jain2025r2egymproceduralenvironmentshybrid} or learnability gap \citep{li2025small}.
Consistently, our ablation analysis in Section \ref{sec:ablation_studies} demonstrates that while the 14B model saturates with more training examples, the 32B model is able to further leverage increased data for performance gains.

\noindent \textbf{AutoSDT-Coder models can handle different types of data-driven coding tasks.} The performance improvement on DiscoveryBench
showcases the generalization capability of models trained on \dataset. In addition to improved performance on the tasks in ScienceAgentBench which give explicit instructions about the type of analysis to be conducted, AutoSDT-Coder models perform better at handling the open-ended hypothesis generation questions in DiscoveryBench.

\input{tables/main_results}

\noindent\textbf{AutoSDT-Coder-32B outperforms larger open-weight models and rivals proprietary models.} As shown in Table \ref{tab:all-models}, we also compare the performance of AutoSDT-Coder models against larger open-weight and proprietary models on ScienceAgentBench. Our models outperform open-weight models of significantly larger size; for example, AutoSDT-Coder-32B achieves more than double the performance of the Llama-3.1-Instruct-405B model. Moreover, AutoSDT-Coder-32B outperforms GPT-4o (2024-05-13) and rivals the performance of Claude-3.5-Sonnet-v1 and GPT-4o (2024-11-20). These results show the effectiveness of \pipeline in generating high-quality scientific data and its potential to train truly open-weight and open-data co-scientist agents. However, as shown in Table \ref{tab:all-models}, we observe that there is still a large gap with reasoning models\footnote{Models that incorporate a "thinking" or "chain-of-thought" process before generating the final answer.} like OpenAI-o1 \citep{openai_o1} and Claude-3.7-Sonnet \citep{anthropic2025claude3.7}. We believe that boosting our data with high-quality reasoning trajectories could be a promising direction to explore and leave it as a future work.

\subsection{Ablation Studies}
\label{sec:ablation_studies}


\input{tables/cross_discipline_results}

\noindent \textbf{Cross-disciplinary Generalization.} 
Table~\ref{tab:cross_discipline_results} shows that \pipeline-Coder-14B achieves decent performance not only on its in-discipline tasks but also generalizes across disciplines to some extent. 
For example, the Bioinformatics model is able to solve problems requiring specialized tools for cell and molecular analysis \citep{wolf2018scanpy, gowers2016mdanalysis}, but is also able to generalize to chemistry tasks due to their shared usage of common scientific libraries and tools. 
Most disciplines do not benefit from adding training data from other domains, thus discipline-specific data collection and training could be a promising direction of future work to build effective specialized models. An exception to this in our experiment is Geographic Information Science, where the specialized model is able to solve domain-specific raster data analysis problems \citep{rasterio},  but training on other disciplines allows it to solve a broader set of problems requiring more general tools. 
Overall, these results suggest that discipline-specific data might be effective in training highly specialized models but multi-discipline training can help one single model tackle a wider range of scientific tasks.

\input{figures/scaling}

\noindent \textbf{Scaling Training Examples.}
We analyze the impact of training set size and model size on ScienceAgentBench in Figure~\ref{fig:scaling}. Specifically, we fine-tune both Qwen2.5-Coder-14B-Instruct and Qwen2.5-Coder-32B-Instruct using 1k, 2.5k, and 5k training examples.
The 14B model exhibits noticeable gains up to 2.5k examples, after which further scaling does not yield improvement, suggesting the onset of performance saturation.
In contrast, the 32B model continues to benefit from additional data, achieving higher success rates as the training set increases to 5k+.
This analysis suggests that performance gains from scaling up training data become limited for smaller models, while larger models are able to better utilize increased data for further improvement.
Such scaling behavior is consistent with previous findings~\citep{jain2025r2egymproceduralenvironmentshybrid}, which report that the 14B model saturates at approximately 800 training trajectories for addressing software engineering issues, while the 32B still benefits from more training trajectories.


\section{Related Work}
\label{app:related}

\noindent\textbf{Scientific Coding Datasets.}  Multiple benchmarks have been proposed to measure the growing capabilities of LLMs in scientific domains such as SciCode \citep{tian2024scicode}, ScienceAgentBench \citep{chen2025scienceagentbench}, DiscoveryBench \citep{majumder2025discoverybench}, BLADE \citep{gu2024bladebenchmarkinglanguagemodel}, and BixBench \citep{mitchener2025bixbenchcomprehensivebenchmarkllmbased}. 
These works rely on human curation of task instances, which is often inefficient and results in small datasets. In contrast, our work is the first to adopt auto-collection, enabling us to create a dataset at a much larger scale.
The closest datasets in scope to \dataset are ScienceAgentBench and DiscoveryBench. Both these benchmarks focus on the assessment of agents' abilities to analyze data and write corresponding code solutions and are grounded in well-defined scientific disciplines. 
We compare \pipeline to related datasets in Table \ref{tab:comparison} in Appendix \ref{app:comparison}.\newline
\noindent\textbf{Automatically Collected Coding Datasets.} Our work is the first to address
the challenging task of collecting data-driven scientific discovery programs at scale. However, multiple automatic approaches have been introduced for software engineering tasks.
RepoST \citep{xie2025repostscalablerepositorylevelcoding} presents a sandboxing approach to build scalable training data for function-level coding. Concurrent to our work, R2E-Gym \citep{jain2025r2egymproceduralenvironmentshybrid} proposes synthetic training instances by backtranslating commits, while SWE-smith \citep{yang2025swesmithscalingdatasoftware} introduces a mostly automatic approach to generate large-scale data for software engineering agents by synthesizing task instances that break the repositories' test cases. 
Our work significantly differs from these by focusing on code for data-driven scientific discovery. In terms of data collection, finding suitable repositories for our purpose requires added effort. In contrast to works that select their repositories from the most popular PyPI packages \citep{yang2025swesmithscalingdatasoftware}, or use SEART GitHub search \footnote{https://seart-ghs.si.usi.ch/} with straightforward criteria such as recency, number of stars, etc. \citep{pan2024trainingsoftwareengineeringagents}, our data collection process requires initial filtering to source suitable repositories followed by file-level checks to ensure the code is related to data-driven coding tasks. Moreover, our focus on well-defined scientific disciplines requires rigorous human evaluation by domain experts to ensure the relevance of the tasks collected through our pipeline. Such evaluation is often unnecessary for more generic software engineering datasets. 
Lastly, collecting coding tasks from research repositories introduces a new set of challenges related to verification since such code does not come with associated unit tests to be leveraged. 
We further discuss the verification challenge in the \hyperref[sec:limitations]{Limitations} section. \newline
\section{Conclusion}
\label{sec:conclusion}
We introduce \pipeline, a fully automatic pipeline to collect data-driven scientific coding tasks at scale. Using \pipeline, we collect \dataset and conduct rigorous evaluation with subject matter experts to confirm the quality of its task instances. We train AutoSDT-Coder-32B, which shows substantial performance gains on two recent challenging data-driven scientific discovery benchmarks. Through \pipeline, we aim to get closer to the ultimate goal of building truly open AI co-scientists.

\section*{Limitations}
\label{sec:limitations}

We recognize the following limitations and future work directions:

\noindent\textbf{Verification of task instances.} \pipeline creates data-driven coding datasets composed of task instructions and code solutions but does not generate evaluation scripts for each code solution, thus limiting its usability in some settings such as reinforcement learning. Unlike software engineering datasets that rely on unit tests that are readily available on popular GitHub repositories, the main challenge in creating evaluation scripts for the tasks in \dataset is that they should be \textit{outcome-based}, which means that the output of the model should be compared against ground-truth results. However, based on our preliminary attempts, it is non-trivial to ensure the complete correctness of the programs adapted through \pipeline without human or even subject matter expert intervention. Relying on alternative methods such as LLM-as-judge or applying heuristics may lead to evaluation scripts that do not measure the true correctness of the program solution and thus be unusable as a signal for reinforcement learning or rejection sampling. An interesting direction of future work would be to implement an automatic and reliable framework to generate instance-specific evaluation scripts which would greatly enhance the use cases of our dataset.

\noindent\textbf{Reasoning Models and Agent Frameworks.} In our work, we mainly focus on improving the performance of base models. We do not train reasoning models on \dataset due to the challenges in generating effective long chain-of-thought (CoT) rationales at scale. Some recent works have investigated generating CoT rationales from code by explaining solution programs \citep{li2024distillingalgorithmicreasoningllms}, which presents a promising direction for future research. Likewise, we leave the experimentation with agent frameworks such as OpenHands CodeAct \citep{wang2025openhandsopenplatformai} and self-debug\citep{chen2024self_debug} for future work. 

\noindent\textbf{Dataset Scale}. Due to resource constraints (e.g., API costs and accessibility of human experts), we only crawl 2,993 repositories to generate 5.4k task instances. However, given that there are more repositories on GitHub and PapersWithCode to crawl, future work can use our pipeline to generate even larger datasets.

\noindent\textbf{Discipline and Programming Language Diversity.} In creating \dataset, we focus on four disciplines that have a wealth of open-source code and experts we can easily contact. However, our pipeline can be used to collect data-driven coding datasets for any discipline that hosts research code on GitHub simply by providing discipline-specific seed keywords. Likewise, our pipeline is geared toward Python since it is the most common programming language in the disciplines of interest. However, \pipeline can be easily extended to other languages commonly used in data analysis such as R and Stata. 

\section*{Ethical Considerations}
\label{sec:ethics}
\pipeline creates tasks based on open-source code and data, and we respect the creators' ownership and intellectual property. We have made our best effort to ensure that the repositories included in \dataset have permissive licenses allowing for academic use. We provide more details in Appendix \ref{app:licenses}.

\section*{Author Contributions}
Y. Li co-led the project, designed and implemented the \pipeline pipeline, ran \pipeline to create \dataset, implemented and conducted all experiments on ScienceAgentBench and DiscoveryBench, and wrote Section \ref{sec:experiments} of the manuscript.
H. N. Moussa co-led the project, designed and implemented the repository searching in \pipeline-Search, contributed to testing and optimizing \pipeline-Select and \pipeline-Adapt, jointly ran \pipeline to collect \dataset, coordinated the expert evaluation, and wrote Sections \ref{sec:pipeline}, \ref{sec:dataset}, \ref{app:related}, \ref{sec:conclusion} of the manuscript.
Z. Chen contributed to the core ideas of the project, gave suggestions to optimize experimental designs, and drafted the abstract and section \ref{sec:intro} of the manuscript.
S. Chen provided feedback on the core ideas and experimental details and revised the manuscript.
B. Yu helped run evaluation experiments on ScienceAgentBench.
M. Xue, B. Burns, T. Chiu, V. Dey, Z. Lu, C. Wei, Q. Zhang, T. Zhang, S. Gao are subject matter experts who validated 256 randomly sampled tasks.
X. Huang and X. Ning are senior subject matter experts in this project. N. Ahmed and A. Payani are senior authors who provided constructive feedback on the project during bi-weekly discussions.
H. Sun is the senior lead author who conceived and oversaw the research project, contributed to the core ideas, and revised the manuscript. 

\section*{Acknowledgements}
The authors would like to thank colleagues from the OSU NLP group and the Amazon AGI team for constructive feedback. The authors thank Frazier N. Baker for participating in a preliminary round of expert evaluation. This research was sponsored in part by NSF OAC 2112606, Amazon, Cisco, and Ohio Supercomputer Center \citep{OhioSupercomputerCenter1987}. The views and conclusions contained herein are those of the authors and should not be interpreted as representing the official policies, either expressed or implied, of the U.S. government. The U.S. Government is authorized to reproduce and distribute reprints for Government purposes notwithstanding any copyright notice herein.

\bibliography{custom}  

\newpage
\appendix
\input{appendix}

\end{document}

%% file: math_commands.tex
\usepackage{amsmath,amsfonts,bm}

\def\eqref#1{equation~\ref{#1}}

\def\1{\bm{1}}

\DeclareMathAlphabet{\mathsfit}{\encodingdefault}{\sfdefault}{m}{sl}
\SetMathAlphabet{\mathsfit}{bold}{\encodingdefault}{\sfdefault}{bx}{n}



%% file: figures/highlight.tex
\begin{figure}[t]
    \centering
    \includegraphics[width=0.98\columnwidth]{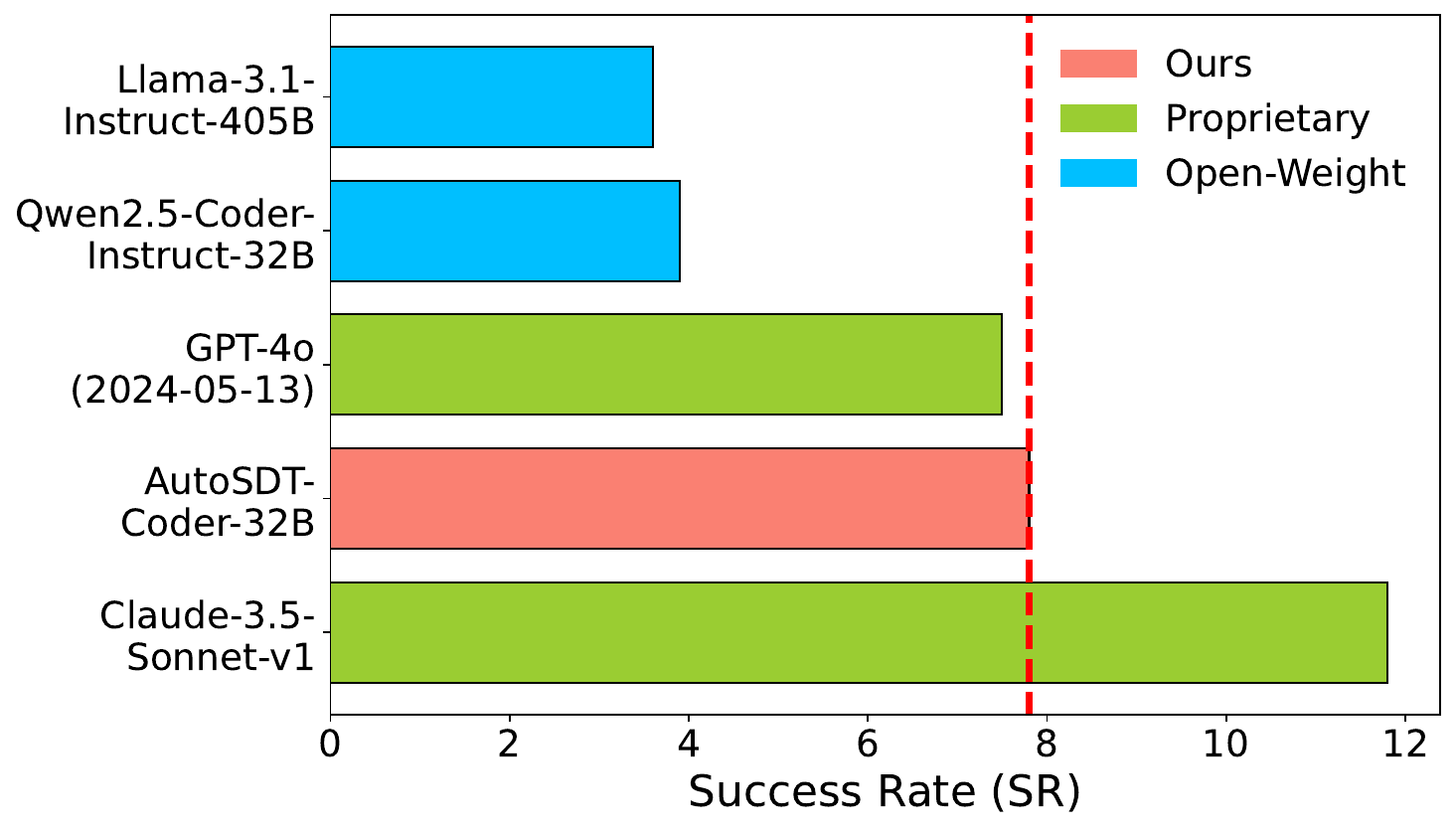}
    \vspace{-11pt}
    \caption{
    \label{fig:highlight}
    Performance of our AutoSDT-Coder in comparison to other open-weight and proprietary LLMs on ScienceAgentBench. AutoSDT-Coder-32B achieves the same level of performance as GPT-4o (2024-05-13).
    }
    \vspace{-16.5pt}
\end{figure}

%% file: figures/AutoDCT.tex
\begin{figure*}[t]
    \centering
    \includegraphics[width=0.92\linewidth]{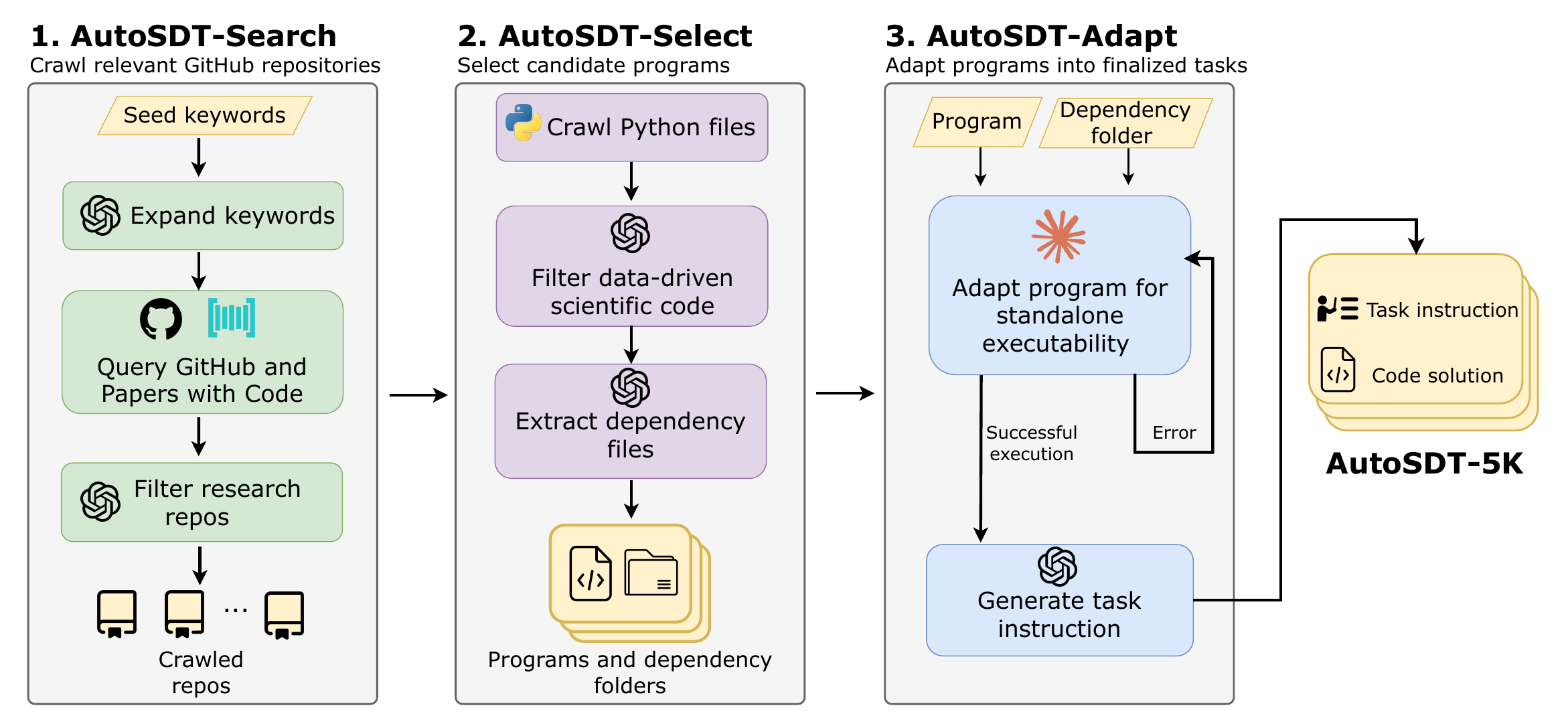}
    \caption{\pipeline collects data-driven discovery tasks in three steps: (1) \textbf{\pipeline-Search} takes a list of keywords for each discipline and searches for relevant repositories. (2) \textbf{\pipeline-Select} identifies programs that represent data-driven discovery tasks and extracts their execution dependency folders. (3) \textbf{\pipeline-Adapt} modifies the selected programs to be independently executable and generates their corresponding task instructions.}
    \vspace{-6pt}
    \label{fig:AutoDCT}
\end{figure*}

%% file: tables/dataset_statistics.tex
\begin{table}[!t]
\small
\centering
\resizebox{0.7\columnwidth}{!}{
\begin{tabular}{lc}
\toprule
\textbf{Statistics} & \textbf{Value} \\
\midrule
\# Tasks & 5,404 \\
\# Repositories & 1,325 \\
\# Packages & 756 \\
Cost (USD) & 2,955 \\
\midrule
\multicolumn{2}{l}{\textbf{Disciplines (\# Tasks/ \# Repositories):}} \\
Bioinformatics & 1,466 / 396 \\
Computational Chemistry & 1,345 / 311 \\
Geo. Info. Science & 1,541 / 341 \\
Psy. \& Cog. Neuroscience & 1,052 / 277 \\
\midrule
Avg \# Tasks/Repo & 3.8 \\ 
Avg \# Subtasks/Task & 4.3 \\ 
Avg \# of lines & 262.8 \\
\bottomrule
\end{tabular}
}
\caption{Detailed statistics of \dataset.}
\label{tab:dataset_statistics}
\vspace{-11pt}
\end{table}

%% file: figures/subtasks_and_packages.tex
\begin{figure*}[t]
    \centering
    \begin{minipage}{0.482\textwidth} 
        \centering
        \includegraphics[width=\linewidth]{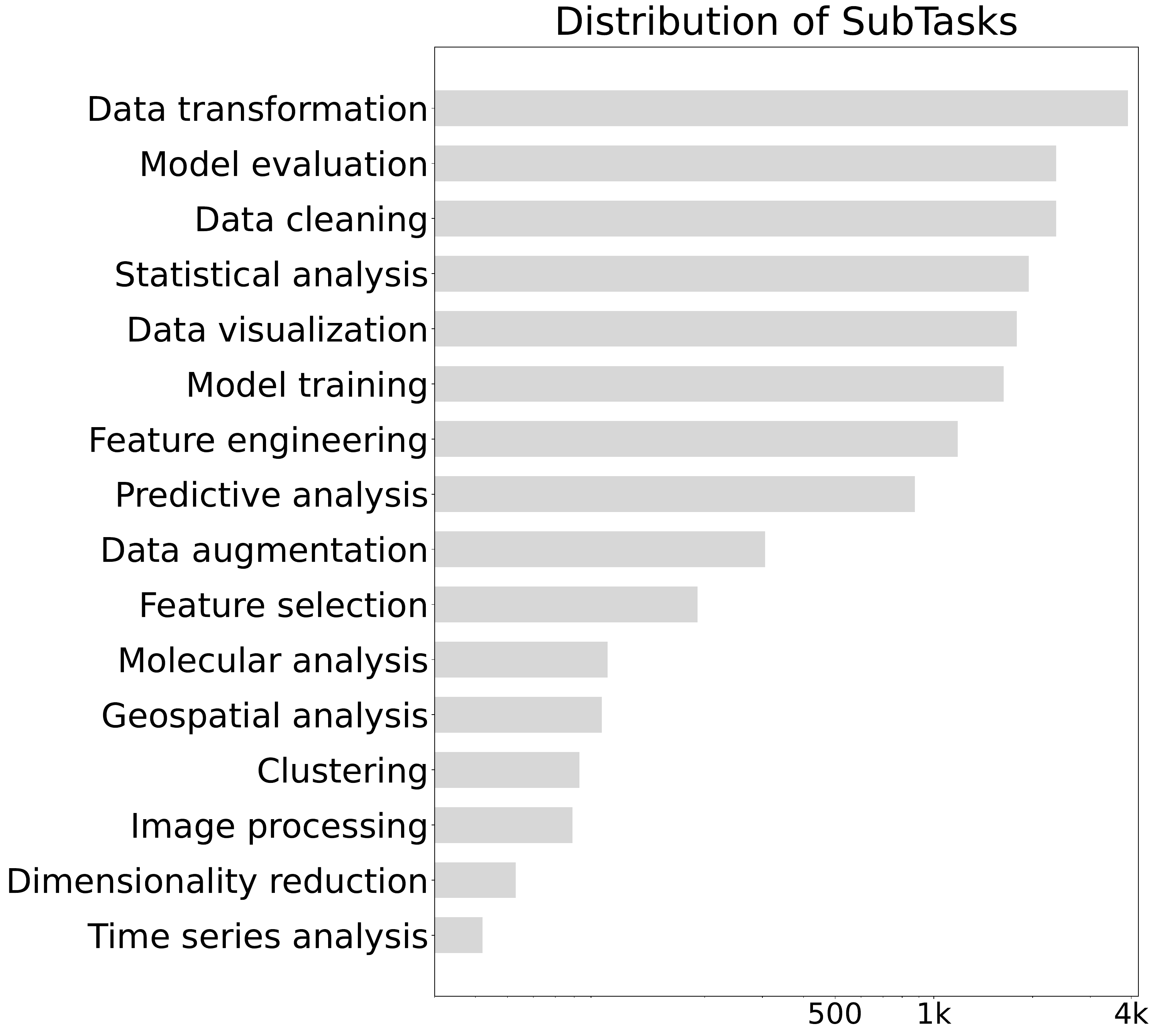} 
        \subcaption{}
        \label{fig:subtasks}
    \end{minipage}\hfill
    \hspace{1mm}
    \begin{minipage}{0.482\textwidth} 
        \centering
        \includegraphics[width=\linewidth]{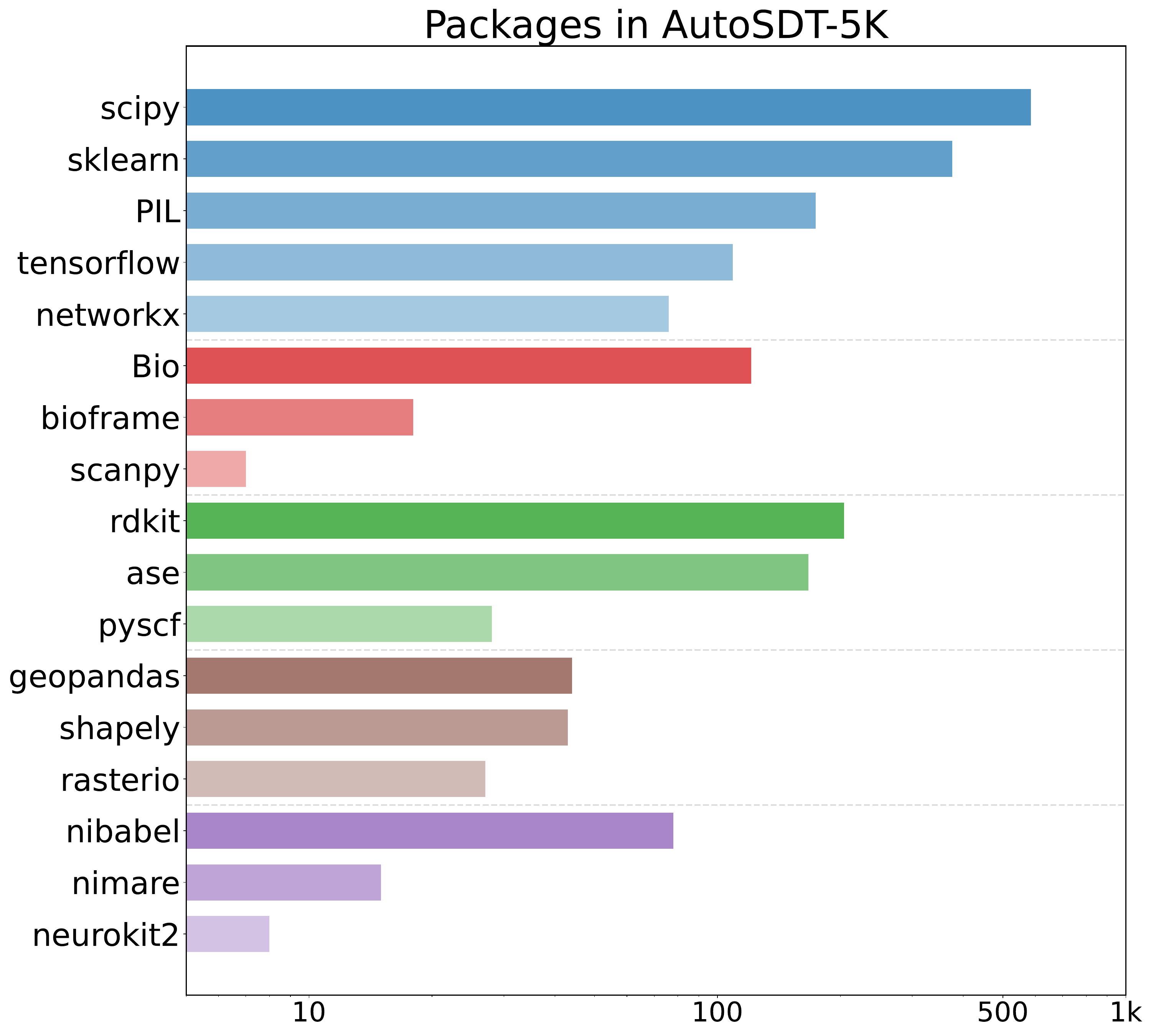} 
        \subcaption{}
        \label{fig:packages}
    \end{minipage}
    \caption{(a) Distribution of subtasks in our dataset. Tasks in \dataset are multi-step research workflows covering operations ranging from data preprocessing to more advanced analytics. (b) Examples of packages in \dataset, including \textcolor[HTML]{4288b9}{\textbf{general-purpose toolkits}} and domain-specific packages for \textcolor[HTML]{d8484b}{\textbf{bioinformatics}}, \textcolor[HTML]{51a953}{\textbf{computational chemistry}}, \textcolor[HTML]{986d65}{\textbf{geographic information science}}, and \textcolor[HTML]{9d7bc0}{\textbf{psychology and cognitive neuroscience}}.
    }
    \label{fig:combined}
    \vspace{-13pt}
\end{figure*}

%% file: tables/analysis_per_difficulty_level.tex
\begin{table}[t]
\small
\centering
\resizebox{0.9\columnwidth}{!}{
\setlength{\tabcolsep}{4pt}
\begin{tabular}{lccc}
\toprule
\textbf{Difficulty} & \textbf{\%} & \textbf{Avg \# of Lines} & \textbf{Avg \# of Subtasks} \\
\midrule
Easy & 22.3 & 214.7 & 4.1 \\
Medium & 48.4 & 263.7 & 4.4 \\
Hard & 29.3 & 403.2 & 5.1 \\
\bottomrule
\end{tabular}
}
\caption{Expert-rated task difficulty distribution and average lines of code and number of subtasks per task for each difficulty level.}
\label{tab:analysis-difficulty-levels}
\vspace{-14pt}
\end{table}

%% file: tables/main_results_sab_v2_expanded_nocbs.tex
\begin{table*}[t]
\small
\centering
\resizebox{1.0\textwidth}{!}{
\begin{tabular}{lccccccccc}

\toprule

\multirow{4}{*}{\makecell[r]{\textbf{Model Size}}} & \multicolumn{6}{c}{\textbf{ScienceAgentBench}} & \multicolumn{3}{c}{\textbf{DiscoveryBench}} \\
\cmidrule{2-10}

 & \multicolumn{3}{c}{\textbf{\textbf{SR}(\%, $\uparrow$)}} & \multicolumn{3}{c}{\textbf{VER} (\%, $\uparrow$)} & \multicolumn{3}{c}{\textbf{HMS} (\%, $\uparrow$)} \\
 \cmidrule{2-10}
 & Base & SFT & $\Delta$ & Base & SFT & $\Delta$ & Base & SFT & $\Delta$ \\ 
\midrule


7B & 3.3 ($\pm$0.5) & 2.3 ($\pm$0.9) & \myred{-1.0 (30\%)} & 19.9 ($\pm$0.2) & 27.5 ($\pm$3.3) & \mygreen{+7.6 (38\%)} & 4.8 ($\pm$1.0) & 6.3 ($\pm$1.3) & \mygreen{+1.5 (31\%)} \\

14B & 4.3 ($\pm$0.5) & 5.9 ($\pm$1.6) & \mygreen{+1.6 (37\%)} & 26.5 ($\pm$2.1) & 35.0 ($\pm$2.5) & \mygreen{+8.5 (32\%)} & 6.4 ($\pm$0.2) & 7.3 ($\pm$0.3) & \mygreen{+0.9 (14\%)} \\

32B & 3.9 ($\pm$0.8) & \textbf{7.8} ($\pm$1.4) & \mygreen{+3.9 (100\%)} & 28.4 ($\pm$0.8) & \textbf{36.0} ($\pm$5.3) & \mygreen{+7.6 (27\%)} & 6.9 ($\pm$0.6) & \textbf{8.1} ($\pm$0.7) & \mygreen{+1.2 (17\%)} \\

\bottomrule
\end{tabular}
}
\vspace{-6pt}
\caption{Results of different-sized Qwen2.5-Coder-Instruct models on ScienceAgentBench and DiscoveryBench. In general, fine-tuned models show substantial performance gains compared with base models. All results are generated with zero-shot direct prompting.}
\label{tab:direct_prompting_results}
\vspace{-12pt}
\end{table*}

%% file: tables/main_results.tex
\begin{table}[t]
\small
\centering
\resizebox{0.999\columnwidth}{!}{
\begin{tabular}{lcc}
\toprule
\textbf{Models} & \textbf{SR (\%, $\uparrow$)} & \textbf{VER (\%, $\uparrow$)} \\
\midrule

\multicolumn{3}{c}{\textit{Proprietary Reasoning Models}} \\
\midrule
Claude-3.7-Sonnet & 18.6 ($\pm$0.8) & 51.6 ($\pm$4.7) \\
OpenAI o1-preview & \textbf{23.9} ($\pm$0.5) & \textbf{56.2} ($\pm$1.7) \\
\midrule

\multicolumn{3}{c}{\textit{Proprietary Non-Reasoning Models}} \\
\midrule
GPT-4o (2024-05-13) & 7.5 ($\pm$0.5) & 42.2 ($\pm$1.6) \\
GPT-4o (2024-11-20) & 11.4 ($\pm$1.2) & 43.1 ($\pm$2.1) \\
Claude-3.5-Sonnet-v1 & 11.8 ($\pm$2.1) & 36.0 ($\pm$1.2) \\
\midrule

\multicolumn{3}{c}{\textit{Open-Weight Models}} \\
\midrule
Llama-3.1-Instruct-70B & 3.6 ($\pm$2.0) & 22.2 ($\pm$0.9) \\
Llama-3.1-Instruct-405B & 3.6 ($\pm$0.5) & 32.0 ($\pm$0.5) \\
Qwen2.5-Coder-Instruct-32B & 3.9 ($\pm$0.8) & 28.4 ($\pm$0.8) \\
\midrule

\multicolumn{3}{c}{\textit{Fine-tuned Open-Weight Models (Ours)}} \\
\midrule
\pipeline-Coder-7B & 2.3 ($\pm$1.2) & 27.5 ($\pm$3.3) \\
\pipeline-Coder-14B & 5.9 ($\pm$1.6) & 35.0 ($\pm$2.5) \\
\pipeline-Coder-32B & 7.8 ($\pm$1.4) & 36.0 ($\pm$5.2) \\
\bottomrule
\end{tabular}
}
\caption{Performance comparison among models on ScienceAgentBench.}
\label{tab:all-models}
\end{table}

%% file: tables/cross_discipline_results.tex
\begin{table}[t]
\small
\centering
\resizebox{0.999\columnwidth}{!}{
\begin{tabular}{lcccc}
\toprule
\textbf{Training Data} & \textbf{Bio.} & \textbf{Chem.} & \textbf{Geo.} & \textbf{Psy \& Neu} \\
\midrule
Bio-only & \textbf{18.5} & 10.0 & 0.0 & \textbf{7.1} \\
Chem-only & 11.1 & \textbf{15.0} & 0.0 & \textbf{7.1} \\
Geo-only & 14.8 & \textbf{15.0} & 3.7 & \textbf{7.1} \\
Psy \& Neu & 11.1 & 5.0 & 3.7 & \textbf{7.1} \\
Full & 11.1 & \textbf{15.0} & \textbf{14.8} & \textbf{7.1} \\
\bottomrule
\end{tabular}
}
\caption{Cross-disciplinary generalization results (SR \%) of the 14B model on ScienceAgentBench. Each row indicates the discipline-only training data, while each column reflects performance measured on ScienceAgentBench specific to the target discipline. The SR is computed by counting a case as successful if at least one out of three independent runs is successful.}
\label{tab:cross_discipline_results}
\vspace{-6pt}
\end{table}

%% file: figures/scaling.tex
\begin{figure}[t]
    \centering
    \includegraphics[width=0.98\columnwidth]   {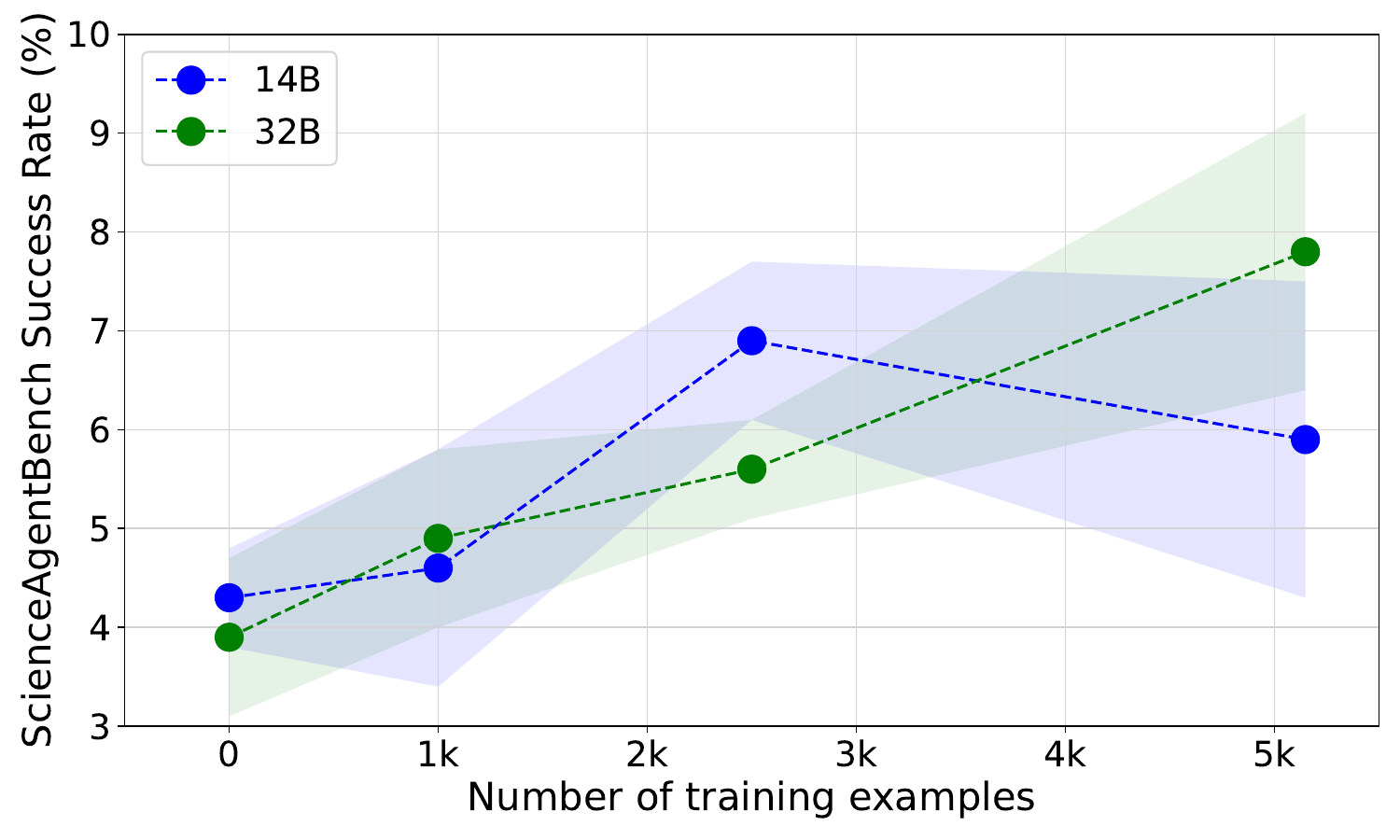}
    \vspace{-11pt}
    \caption{Impact of training set size on ScienceAgentBench performance for AutoSDT-Coder-14B and AutoSDT-Coder-32B. Shaded areas indicate standard deviation across three runs.}
    \label{fig:scaling}
\end{figure}

%% file: appendix.tex
\section*{Appendix}
We provide more details omitted from the main text in the Appendix as follows:
\begin{itemize}
    \item Appendix \ref{app:pipeline-details}: Details and Prompts of \pipeline
        \begin{itemize}
            \item Appendix \ref{app:crawl}: \pipeline-Search
            \item Appendix \ref{app:select}: \pipeline-Select
            \item Appendix \ref{app:adapt}: \pipeline-Adapt \\
        \end{itemize}
    \item Appendix \ref{app:examples}: Example Tasks
        \begin{itemize}
            \item Appendix \ref{app:instructions}: Task Instructions
            \item Appendix \ref{app:full-example}: Full Task Example \\
        \end{itemize}
    \item Appendix \ref{app:cost}: Cost Breakdown \newline
    \item Appendix \ref{app:eval}: Expert Evaluation \newline
    \item Appendix \ref{app:training-details}: Training Details \newline
    \item Appendix \ref{app:comparison}: Related Datasets \newline
    \item Appendix \ref{app:licenses}: Repository Licenses \newline
\end{itemize}

\setcounter{table}{0}
\renewcommand\thetable{\Alph{section}.\arabic{table}}
\setcounter{figure}{0}
\renewcommand\thefigure{\Alph{section}.\arabic{figure}}
\setcounter{table}{0}
\renewcommand\thetable{\Alph{section}.\arabic{table}}
\setcounter{figure}{0}
\renewcommand\thefigure{\Alph{section}.\arabic{figure}}
\section{Details and Prompts of \pipeline}
\label{app:pipeline-details}
\subsection{\pipeline-Search}
\label{app:crawl}

\input{tables/comparison}
\input{tables/crawl_prompt}
In order to search for suitable repositories we use the GitHub GraphQL API\footnote{https://docs.github.com/en/graphql} and the PapersWithCode API\footnote{https://paperswithcode.com/api/v1/docs/}. 

After seed keyword expansion using GPT-4o, we obtain the following expanded keywords per discipline: bioinformatics (\textit{genomics, biomarkers, proteomics}), computational chemistry (\textit{molecular dynamics, cheminformatics, catalysis}), psychology (\textit{psychometrics, neuropsychology, cognition}), neuroscience (\textit{neuroimaging, neuroplasticity, neuroinformatics}), geographic information science (\textit{geoscience, geospatial, cartography}). 

The seed and expanded keywords are then used to query the GitHub GraphQL API, which searches for repositories containing these keywords within their README.md files or descriptions, alongside terms commonly indicative of research-oriented repositories (e.g., ``citation,'' ``doi,'' and ``arxiv''). In order to control the quality of retrieved repositories, we restrict results to Python-based repositories with a minimum of 10 stars.

After a repository is identified via GitHub or PapersWithCode, it goes through an LLM filtering stage using GPT-4o which checks that the repository hosts research code related to the discipline of interest and if so extracts the links to the papers. The prompt used for filtering is given in Table \ref{tab:crawl-prompt}. For PapersWithCode the link extraction stage is skipped since the arXiv paper links can be obtained directly from the API. 
\subsection{\pipeline-Select}
\label{app:select}
\input{tables/scientific_task_verify_prompt}
\input{tables/locate_dependencies_prompt}
After identifying suitable repositories using \pipeline-Search, we locally clone them for further processing. This is because we would quickly reach the GitHub GraphQL rate limit if we perform these operations via API. Once the repositories are cloned and all Python files are identified, we first perform rule-based filtering to eliminate files that are excessively lengthy (i.e., more than 1000 lines) and those located in directories unlikely to contain substantive scientific programs (e.g., ``utils,'' ``config,'' ``tests''). The remaining files then undergo LLM-based filtering using GPT-4o to judge whether they host code for data-driven scientific discovery. The prompt containing the detailed criteria is given in Table \ref{tab:scitaskverify}. 

In order to locate dependencies, we use GPT-4o with the prompt given in Table \ref{tab:locate}. The LLM is given both the code and the file structure of the repository and is asked to return the paths of the dependencies contained within the repository. These can be dataset files, models, local modules, etc. 
\subsection{\pipeline-Adapt}
\label{app:adapt}
In the program adaptation stage of \pipeline-Adapt, we prompt Claude-3.7-Sonnet with the original program and the structure of the dependency folder and explicitly instruct it to only make minimal changes required to ensure the executability of the program and not alter the original functionality. The prompt that we use is given in Table \ref{tab:adapt}. In order to generate instructions, we use GPT-4o with the prompt given in Table \ref{tab:instructiongen}.
\input{tables/adapt_prompt}
\input{tables/instruction_prompt}
\setcounter{table}{0}
\renewcommand\thetable{\Alph{section}.\arabic{table}}
\setcounter{figure}{0}
\renewcommand\thefigure{\Alph{section}.\arabic{figure}}
\input{tables/cost}
\input{tables/main_results_db}
\section{Example Tasks}
\label{app:examples}
\subsection{Task Instructions}
\label{app:instructions}
We provide examples of task instructions for each of the disciplines covered in \dataset in Table \ref{tab:examples}. 
\input{tables/examples}
\subsection{Full Task Example}
\label{app:full-example}
We provide an example of a (task instruction, code solution) pair in Geographic Information in Listing \ref{list:full-example-p1}.

\input{geo_example}
\setcounter{table}{0}
\renewcommand\thetable{\Alph{section}.\arabic{table}}
\setcounter{figure}{0}
\renewcommand\thefigure{\Alph{section}.\arabic{figure}}
\section{Cost Breakdown}
\label{app:cost}
We show the detailed breakdown of the API cost for each stage of \pipeline in order to build \dataset. For \pipeline-Search, \pipeline-Select, and the instruction generation in \pipeline-Adapt, we use GPT-4o. For code adaptation in \pipeline-Adapt we use Claude-3.7-Sonnet.

\setcounter{table}{0}
\renewcommand\thetable{\Alph{section}.\arabic{table}}
\setcounter{figure}{0}
\renewcommand\thefigure{\Alph{section}.\arabic{figure}}
\input{tables/licenses}
\section{Expert Evaluation}
\label{app:eval}
\input{expert_questionnaire}

\setcounter{table}{0}
\renewcommand\thetable{\Alph{section}.\arabic{table}}
\setcounter{figure}{0}
\renewcommand\thefigure{\Alph{section}.\arabic{figure}}
\section{Training Details}
\label{app:training-details}
\paragraph{Supervised Fine-tuning.}
We perform full parameter fine-tuning using the LlamaFactory library \citep{zheng2024llamafactory} . For AutoSDT-Coder-7B/14B/32B, we train them with learning rate 1e-5, maximum 1 epoch, and a max context length of 8192.
Warmup is turned off for 7B/14B and turned on for 32B.
Training is done on 4 NVIDIA H100 96G GPUs (for 7B/14B) and 8 for 32B models.

\paragraph{Inference.} We use the vLLM library \citep{kwon2023efficient} to deploy LLM servers and conduct inference experiments.
For all the inference in ScienceAgentBench, we use a default temperature=0.2, top\_p=0.95, and max\_tokens=2000. For the inference in DiscoveryBench, we use a default temperature=0.2, top\_p=0.95, and max\_tokens=1024.
\paragraph{Re-implementation of Inference of DiscoveryBench.} We re-implemented the inference pipeline based on the original codebase provided by DiscoveryBench authors. Their original implementation was based on LangChain to build an end-to-end LLM agent, while our need is to decouple the code generation step. Therefore, the results of DiscoveryBench in this paper are based on our reproduction and are slightly different from the original paper. We compare our reproduced results and the numbers in the DiscoveryBench paper in Table \ref{table:result_db}.

\section{Related Datasets}
\label{app:related_datasets}
We compare \dataset against existing science-oriented data analysis datasets in Table \ref{tab:comparison}.
(1) \dataset covers a considerably larger set of tasks balanced across multiple disciplines. 
(2) Unlike MLE-Bench and DSBench which derive code from competition platforms or RE-Bench and Bix-Bench which use human annotators to curate new coding tasks, \dataset is based on naturally-occurring code authored by real-world scientists, ensuring the ecological validity of the tasks. 
(3) \dataset is the only automatically generated dataset for coding tasks in scientific disciplines.
While automatic generation approaches have been applied to software engineering datasets \cite{xie2025repostscalablerepositorylevelcoding,xie2024codebenchgencreatingscalableexecutionbased,yang2025swesmithscalingdatasoftware}, 
this work is the first to address
the challenging task of collecting high-quality data-driven scientific discovery programs at scale. 
\label{app:comparison}

\section{Repository Licenses}
\label{app:licenses}
We ensure that all 1325 repositories composing the final tasks in \dataset allow for academic use. We list the licenses and the number of corresponding repositories in Table \ref{tab:licenses}. We manually checked the 15 repositories with custom licenses and ensured that they all allow academic and non-commercial use and list them in Table \ref{tab:other-licenses}. There are also 317 repositories without any license information. We assume that these repositories are permissive for academic purposes.

%% file: tables/comparison.tex
\begin{table*}[t]
\small
\centering
\vspace{-5.5pt}
\begin{tabular}{@{\hspace{0pt}}lccccc@{\hspace{0pt}}}
\toprule
\multirow{2}{*}{\textbf{Dataset}} & \textbf{Task} & \textbf{Subject} & \textbf{Scientific} & \textbf{Naturally} & \textbf{Auto} \\
\textbf{} & \textbf{Instances} & \textbf{Domains} & \textbf{Dataset} & \textbf{Occurring Code} & \textbf{Collection}\\
\midrule
DA-Code \citep{huang2024dacodeagentdatascience} & 500 & 0 & \xmark & \cmark & \xmark \\
DSBench \citep{jing2025dsbenchfardatascience} & 540 & 0 & \xmark & \xmark & \xmark \\
MLE Bench \citep{chan2025mlebenchevaluatingmachinelearning} & 75 & 1 & \xmark & \xmark & \xmark \\
REBench \citep{wijk2024rebenchevaluatingfrontierai} & 7 & 1 & \xmark & \xmark & \xmark \\
ScienceAgentBench \citep{chen2025scienceagentbench} & 102 & 4 & \cmark & \cmark & \xmark \\
DiscoveryBench \citep{majumder2025discoverybench} & 239 & 6 & \cmark & \cmark & \xmark \\
BLADE \citep{gu2024bladebenchmarkinglanguagemodel} & 12 & 6 & \cmark & \cmark & \xmark \\
BixBench \citep{mitchener2025bixbenchcomprehensivebenchmarkllmbased} & 296 & 1 & \cmark & \xmark & \xmark \\
\midrule
\textbf{\dataset (Ours)} & 5404 & 4 & \cmark & \cmark & \cmark \\
\bottomrule
\end{tabular}
\caption{Dataset statistics of \dataset compared to related datasets. Columns show the number of instances, number of subject domains, and whether the dataset is oriented towards scientific disciplines, based on naturally-occurring code, and automatically collected.}
\label{tab:comparison}
\end{table*}

%% file: tables/crawl_prompt.tex
\begin{table*}[htbp]
\small
\centering
\begin{tabular}{p{0.9\linewidth}}
\toprule
You are an expert at reading GitHub \texttt{README.md} files thoroughly and determining whether the repository hosts code related to a research paper or not, and you are also skilled at correctly extracting the link to the related paper.\newline

Your answer should be based on your thorough understanding of the content of the \texttt{README.md} file. Does the \texttt{README.md} file indicate that the repository hosts code related to a research paper in the discipline of \texttt{\{keyword\}}? Answer by `YES' or `NO' in the `RESEARCH'. If your answer to the previous question is `YES', extract the link to the related research paper. Make sure to extract the link to the research paper that this repository implements only, this should  be the link to the paper that people would cite if they used the code in the repository for their work, ignoring all other irrelevant links that might be referenced in the \texttt{README.md} file. Put the link(s) in front of the `LINKS': as a list of links.\newline

\texttt{README.md} file: \texttt{\{readme\}}\newline

You should strictly follow the format below: \newline

RESEARCH: 

LINKS: \\
\bottomrule
\end{tabular}
\caption{\label{tab:crawl-prompt}
Prompt for AutoSDT-Search Repository Filtering Stage.
}
\end{table*}

%% file: tables/scientific_task_verify_prompt.tex
\begin{table*}[htbp]
\small
\centering
\begin{tabular}{p{0.9\linewidth}}
\toprule
You are an expert at determining whether a program contains scientific code or not. Given a code file, you need to verify if the current code is a scientific task. Several conditions should be satisfied: \newline

1. Functionality: the functionality of the given program should be related to tasks in a scientific workflow. These tasks include but are not limited to feature engineering, machine learning, deep learning, computational analysis, data visualization, model training, numerical calculation/analysis, statistical methods, domain-specific analysis/simulation, etc. \newline
2. Input: the program should receive at least one or multiple datasets as input. In other words, the program is dealing with a dataset and conducting analysis or experiments on top of the data. The data can either be loaded through built-in functions or be loaded from local files. If the current program does not receive and process any data, it cannot be considered as ``a scientific task'' here.\newline
3. Output: the program should output numerical or visualization results that can be further evaluated. \newline

A code file is considered a scientific task ONLY IF it completely satisfied the three dimensions above. For example, code files that purely contain modeling, training/testing, data pre-processing, or only consist of utility functions or class definitions, are not considered a scientific task.\newline

Program name: \texttt{\{file\_name\}} \newline
Program code: \texttt{\{code\}} \newline

After reasoning about the problem, output your final answer strictly based on the following format: \newline
VERDICT: \texttt{\{YES/NO\}} \\
\bottomrule
\end{tabular}
\caption{\label{tab:scitaskverify}
Prompt to Verify Data-driven Scientific Discovery Code.
}
\end{table*}

%% file: tables/locate_dependencies_prompt.tex
\begin{table*}[htbp]
\small
\centering
\begin{tabular}{p{0.9\linewidth}}
\toprule
You are an expert software engineer who is very skilled at analyzing Python code files and their repositories to extract dependencies. \newline

In this task you will be given a Python file and the GitHub file tree of the repository it belongs to, your job is to thoroughly understand the code and all the in-repository dependencies it needs. This is because we would like to run this code in a standalone environment and we have to make sure that all the dependencies that the code needs are copied in that environment. Hence, it is very important that you have a thorough understanding of the code and extract all in-repository dependencies needed. \newline

Specifically, your job is to do the following: \newline 

1. Recognize whether the code makes use of a dataset. The dataset can either be loaded via built-in library functions (e.g., data = MNIST ()) or loaded from a local file in the repository (csv, jsonl, xls, txt, parquet, or any other file type). If the dataset(s) used in the code are either loaded through built-in library functions or contained within the repository, you should output ``Yes'' in DATASET\_LABEL field.  Otherwise, you should output ``No''. \newline

2. In the case where the dataset used in the code is contained within the repository, you also have to find the relative path to the dataset file, based on the GitHub file tree that will be given to you. You will list the paths to all datasets used in the code as a list of paths after the field DATASET\_PATHS. \newline

3. Besides the dataset, now you have to identify all other in-repository dependencies that the code uses, and extract their relative paths based on the file tree given to you. These can be modules, classes, models, or any other dependency that the code imports from a folder within the repository. If you identify that there are in-repository dependencies used, you should put a ``Yes'' in the MODULE\_LABEL. Otherwise, output a ``No''. \newline

4. In the case of a ``Yes'', make sure to put the relative paths to all dependencies as a list of paths in the MODULE\_PATHS field, based on the GitHub file tree given to you. \newline

5. If based on the code alone you can only identify the folder that contains the dependency but not the exact file only return the path to the folder. This is because you might sometimes not be able to know which file the dependency is exactly located in based on only looking at the file tree. Thus, to stay on the safe side, just give the path to the folder that contains the dependency. \newline
                    
Python code: \texttt{\{code\}} \newline

Project directory: \texttt{\{directory\}} \\
\bottomrule
\end{tabular}
\caption{\label{tab:locate}
Prompt to Locate Dependencies.
}
\end{table*}

%% file: tables/adapt_prompt.tex
\begin{table*}[htbp]
\small
\centering
\begin{tabular}{p{0.9\linewidth}}
\toprule
You are an excellent coder at adapting existing files for standalone executability. You will be given a code file from a Github Repo. Your task is to modify the code into a self-contained program that can be run locally and separately.\newline

Please do not change the original functionality of the code. You must keep the original logic and functionality of the code as much as possible. You should never include dummy/pass statements or empty/mock functions in your response. \newline

You need to slightly modify the source code's input/output logistics and intermediate steps to make it a stand-alone program that can be executed locally. The modified code will then be executed in a local environment. If there are errors, you need to debug the code based on the execution feedback.
All the datasets and dependency files are located at \texttt{\{dataset\_path\}}. If the original code has imported modules from local files, you can assume they exist and do the same imports in your modified code. Here is the directory structure of the dataset and dependency files:
\texttt{\{dataset\_structure\}}\newline

Make sure that the code you generate uses the same input files as the original code. Do not generate dummy input files or input data. \newline
                                          
For the output of the programs, your code should save the results to a file named \texttt{``pred\_results/pred\_[code\_file\_name].[extension]''}, depending on the type of data such as \texttt{csv}, \texttt{txt}, \texttt{jsonl}, etc. ALL outputs of the program should be saved in the directory \texttt{pred\_results/}. You should never create new folders or files outside of the specified directory. \newline

Code to be modified: \newline

\texttt{\{code\_file\_name\}}\newline

\texttt{\{code\}}\newline

The user may execute your code and report any exceptions and error messages.
You should address the reported issues and respond with a fixed, complete program.
Note that, when addressing bugs, you should ONLY focus on addressing the errors and exceptions and MUST NOT change or delete the main functionality and logic of the original program just to make it executable.\newline

Keep your response concise and do not use a code block if it's not intended to be executed.
Do not suggest a few line changes, incomplete program outline, or partial code that requires the user to modify. Your response should include a complete, standalone, executable program. \newline

Do not use any interactive Python commands in your program, such as `!pip install numpy`, which will cause execution errors.\newline

Regardless of the iterations of self-debugging, make sure to wrap your program in a code block that specifies the script type, python. For example:
```python
print("Hello World!")
'''\\
\bottomrule
\end{tabular}
\caption{\label{tab:adapt}
Prompt for Code Adaptation.
}
\end{table*}

%% file: tables/instruction_prompt.tex
\begin{table*}[htbp]
\small
\centering
\begin{tabular}{p{0.9\linewidth}}
\toprule
You are a helpful agent for generating task instructions based on a code snippet for solving scientific data processing tasks. You need to provide a clear and concise instruction that best describes the functionality of the given code. The instruction should be written in plain English and should be detailed enough so that a person who has no knowledge of the code can understand the task and implement code for it. The instruction should not reveal too many implementation details but also should be precise and not vague. It should be a high-level description of the code's functionality. \newline

You should thoroughly read the scientific data processing code snippet provided, understand the underlying domain-specific concepts behind it, and generate a task instruction that makes correct use of the domain-specific language. In other words, your task instructions should be written as if they are from a domain scientist giving instructions to a junior researcher in their lab. \newline

The structure of the instruction should be as clear as possible: you should clearly specify the goal of the task, clearly name the exact input file/files that should be used, and the output files that should be created and the path to which they should be saved. Additionally, if the output of the program is written to a file, you should specify the format that the output should be written in, based on the implementation given in the code snippet. In cases where, based on your understanding of the code, you deem that the instruction needs more details - for example, if a certain program can use different computational methods to reach a solution - you can add guidelines about the specific method to use in the instruction. In all cases, ensure that the instruction does not include too many implementation details but also that it is precise and does not invite ambiguity or confusion. The format of your instruction should be a concise paragraph of a few lines without any sections. Keep the instruction focused on the high level scientific goal of the task and do not make reference to unnecessary details like "ensure the directory or so and so files exist". Such low level implementation details should never be part of the instruction. \newline

Please generate the instruction based on the code snippet below.\newline

\texttt{\{code\}}\newline \\
\bottomrule
\end{tabular}
\vspace{-5pt}
\caption{\label{tab:instructiongen}
Prompt for Instruction Generation.
}
\end{table*}

%% file: tables/cost.tex

\begin{table}[t]
\small
\centering
\begin{tabular}{lr}
\toprule
\textbf{Stage} & \textbf{Cost (USD)} \\ 
\midrule
\textbf{\pipeline-Search:} & \\
Repository Crawling & 32 \\
\midrule
\textbf{\pipeline-Select:} & \\
Scientific Task Filtering & 459 \\
Dependency Locating & 828 \\
\midrule
\textbf{\pipeline-Adapt:} & \\
Program Adaptation & 1,210 \\
Instruction Generation & 426 \\
\midrule
\textbf{Total Cost} & 2,955 \\
\bottomrule
\end{tabular}
\caption{\pipeline Cost Breakdown.}
\label{tab:cost}
\end{table}

%% file: tables/main_results_db.tex
\begin{table}[htbp]
\centering
\label{tab:direct_prompting_results_db}
\resizebox{0.35\textwidth}{!}{
\begin{tabular}{lc}
\toprule
\textbf{Model Size} & \textbf{HMS(\%)} \\
\midrule
\multicolumn{2}{c}{\textit{Re-implementation Results}} \\
\midrule
Qwen2.5-Coder-7B-Instruct & 4.8 \\
\pipeline-Coder-7B & 6.3 \\
Qwen2.5-Coder-14B-Instruct & 6.4 \\
\pipeline-Coder-14B & 7.3 \\
Qwen2.5-Coder-32B-Instruct & 6.9 \\
\pipeline-Coder-32B & 8.1 \\
GPT-4o (2024-05-13) & 10.4 \\
\midrule
\multicolumn{2}{c}{\textit{Results copied from DiscoveryBench}} \\
\midrule
Llama-3-70B & 12.1 \\
GPT-4o (2024-05-13) & 15.5 \\
\bottomrule
\end{tabular}
}
\caption{Direct Prompting results on DiscoveryBench.}
\label{table:result_db}
\end{table}

%% file: tables/examples.tex
\begin{table*}[htbp]
\small
\centering
\vspace{-5.5pt}
\resizebox{\textwidth}{!}{
\begin{tabular}{p{3.5cm}p{12cm}}
\toprule
\textbf{Discipline} & \textbf{Task Instruction} \\
\midrule
Bioinformatics & 
\textit{Predict circRNA-disease associations using the Random Walk with Restart (RWR) algorithm. Utilize the circRNA-disease association data in ``circrna\_disease.txt'', along with circRNA and disease lists from ``circ\_list.csv'' and ``dis\_list.csv'' respectively. Perform 5-fold cross-validation to evaluate prediction performance, calculating metrics such as accuracy, recall, precision, F1-score, AUC, and AUPR. Save the results to ``RWR.csv'' in CSV format, including metrics and their values. } \\ 
\midrule
Computational Chemistry & 
\textit{Cluster molecular structures based on their chemical fingerprints using the SMILES data in ``smiles.csv''. Compute Morgan fingerprints for the molecules, perform clustering using the Butina algorithm with a similarity cutoff of 0.72, and identify the centroid molecule for each cluster. Save the clustering summary, including the number of clusters and centroid SMILES, to ``clustering.txt'' and generate SVG visualizations of the centroid molecules for each cluster and save them as ``centroid.svg''.} \\ 
\midrule
Geographic Inf. Sci. & 
\textit{Match geo-tagged drone images to corresponding satellite map images using geographic coordinates. Use the satellite map data from ``map.csv'' and the drone photo metadata from ``metadata.csv''. For each drone image, determine its location on the satellite map by comparing its geographic coordinates with the boundaries of the satellite images. Calculate the drone image's precise geographic position within the matched satellite image and compare it to the ground truth coordinates. Save the results, including the calculated coordinates, errors, and matching status, to ``results.csv''.} \\ 
\midrule
Psy. and Cog. Neuroscience & 
\textit{Process MRI data to calculate the incidence sizes of parental brain regions. Use the MRI in mri.nii.gz and the Allen Brain annotation file allen.nii.gz. Apply a threshold to identify stroke-affected regions and generate the following outputs: (1) a labeled NIfTI file highlighting affected regions saved as ``affected\_regions\_parental.nii.gz'', (2) a text file summarizing stroke volume and affected region percentages saved as ``summary.txt'', and (3) a MATLAB file with detailed region labels and metrics saved as ``label\_count.mat''.} \\ 
\bottomrule
\end{tabular}
}
\vspace{-7pt}
\caption{Representative examples of task instructions for each discipline.}
\label{tab:examples}
\end{table*}

%% file: geo_example.tex
\begin{listing*}[t]
\centering
\begin{minipage}{\textwidth}
\textbf{Task Instruction:} Generate binary road masks by creating buffers around road geometries defined in GeoJSON files located in ``geojson\_roads\_speed/''. Use the corresponding satellite imagery files from the ``PS-RGB'' subdirectory to rasterize the buffered road geometries into binary masks. Save the resulting masks as PNG files in the directory specified by ``output\_mask\_path''. Additionally, save a list of unavailable imagery files (those without corresponding GeoJSON labels) to ``pred\_results/pred\_unavailable\_files.txt''. Ensure the buffer distance around roads is set to the value of ``buffer\_meters'' (default: 2 meters), and assign a pixel value of ``burnValue'' (default: 255) for road areas in the masks.\newline

\textbf{Code Solution:} 
\end{minipage}

\begin{lstlisting}[language=Python]
import argparse
import os
import sys
import time
import numpy as np
from tqdm import tqdm
import cv2
import json
import rasterio
from rasterio import features
from shapely.geometry import shape, mapping
import matplotlib.pyplot as plt

# Create pred_results directory if it doesn't exist
if not os.path.exists('pred_results'):
    os.makedirs('pred_results')

def create_buffer_geopandas(geoJsonFileName, bufferDistanceMeters, bufferRoundness=1,
                            projectToUTM=True, verbose=False):
    """
    Create a buffer around the line segments of the geojson file.
    Return a buffered geometry.
    """
    
    # Load geojson file
    try:
        with open(geoJsonFileName, 'r') as f:
            geojson_data = json.load(f)
    except Exception as e:
        print(f"create_buffer_geopandas(): can't load GeoJSON file: {geoJsonFileName}, error: {e}")
        return None
    
    if not geojson_data.get('features'):
        return None
    
    # Extract geometries
    geometries = []
    for feature in geojson_data['features']:
        if feature.get('geometry'):
            geom = shape(feature['geometry'])
            geometries.append(geom)
    
    if not geometries:
        return None
    
    # Create buffers
    buffered_geometries = []
    for geom in geometries:
        buffered_geom = geom.buffer(bufferDistanceMeters / 111000.0, bufferRoundness)  # Approximate conversion from meters to degrees
        buffered_geometries.append(buffered_geom)
    
    return buffered_geometries
\end{lstlisting}
\caption{Road mask generation task and code solution (page 1 of 3)}
\label{list:full-example-p1}
\end{listing*}

\begin{listing*}[t]
\begin{lstlisting}[language=Python]
def get_road_buffer(geoJson, im_file, output_raster, buffer_meters=2, 
            burnValue=150, bufferRoundness=6, 
            plot_file='', figsize=(6, 6), fontsize=8, dpi=500, 
            show_plot=False, verbose=False):
"""
Create buffer around roads defined in geoJson file, then burn values to an 
output_raster
"""

    buffered_geometries = create_buffer_geopandas(geoJson, buffer_meters, 
                                     bufferRoundness=bufferRoundness, 
                                     projectToUTM=True, verbose=verbose)
    
    if not buffered_geometries:
        return None, None
    # Create the mask
    with rasterio.open(im_file) as src:
        src_profile = src.profile
        out_arr = np.zeros((src.height, src.width), dtype=np.uint8)
    
    # Prepare shapes for rasterization
    shapes = [(mapping(geom), burnValue) for geom in buffered_geometries]
    # Burn the shapes into the raster
    with rasterio.open(im_file) as src:
        out_arr = features.rasterize(shapes=shapes, 
                                    out=out_arr,
                                    transform=src.transform)
    # Save the mask if output_raster is provided
    if output_raster:
        # Create output directory if it doesn't exist
        output_dir = os.path.dirname(output_raster)
        if output_dir and not os.path.exists(output_dir):
            os.makedirs(output_dir)
        
        # Write the mask using OpenCV
        cv2.imwrite(output_raster, out_arr)
    
    return out_arr, buffered_geometries
def create_masks(path_data, buffer_meters=2, is_SN3=True,
                 burnValue=150, make_plots=True, overwrite_ims=False,
                 output_mask_path='',
                 header=['name', 'im_file', 'im_vis_file', 'mask_file',
                         'mask_vis_file']):
    t0 = time.time()
    # set paths
    path_labels = os.path.join(path_data, 'geojson_roads_speed/')
    # output directories
    path_masks = output_mask_path
    # image directories
    if is_SN3:
        #old directory RGB-PanSharpen-u8
        path_images_vis = os.path.join(path_data, 'PS-RGB')
    else:
        path_images_vis = os.path.join(path_data, 'PS-RGB') \
            if os.path.isdir(os.path.join(path_data, 'PS-RGB')) else os.path.join(path_data, 'PS-RGB-u8')
    
    # Create output directory if it doesn't exist
    if not os.path.exists(path_masks):
        os.makedirs(path_masks)
        
    outfile_list = []
    
    # Check if the image directory exists
    if not os.path.exists(path_images_vis):
        print(f"Image directory does not exist: {path_images_vis}")
        return
    
    im_files = os.listdir(path_images_vis)
    nfiles = len(im_files)
    unavailabel = []
    
    for i, im_name in enumerate(tqdm(im_files)):
        if not im_name.endswith('.tif'):
            continue

        # define files
        name_root = os.path.basename(im_name)
        im_file_vis = os.path.join(path_images_vis, im_name)
       
        lab_name = name_root.replace('.tif', '.geojson').split('_')

\end{lstlisting}
\caption{Road mask generation code solution (continued, page 2 of 3)}
\label{list:full-example-p2}
\end{listing*}

\begin{listing*}[t]
\begin{lstlisting}[language=Python]
        label_file = os.path.join(path_labels,
                                  ''.join(['_'.join(lab_name[0:-1]), '_geojson_roads_speed_', lab_name[-1]]))
        label_file_tot = label_file.replace('PS-RGB_', '')
        
        if os.path.isfile(label_file_tot):
            mask_file = os.path.join(path_masks, name_root.replace('.tif', '.png'))
            if not os.path.exists(mask_file) or overwrite_ims:
                try:
                    mask, gdf_buffer = get_road_buffer(label_file_tot,
                                                      im_file_vis,
                                                      mask_file,
                                                      buffer_meters=buffer_meters,
                                                      burnValue=burnValue,
                                                      bufferRoundness=6,
                                                      plot_file='',
                                                      figsize=(6, 6),
                                                      fontsize=8,
                                                      dpi=500,
                                                      show_plot=False,
                                                      verbose=False)
                
                    if mask is not None:
                        cv2.imwrite(mask_file, mask)
            except Exception as e:
                print(f"Error processing {im_name}: {e}")
            else:
                unavailabel.append(im_name)
    print(len(unavailabel), '  Unavilable out of ', nfiles)
    t4 = time.time()
    print("Time to run create_masks():", t4 - t0, "seconds")
# Save list of unavailable files to output
    with open('pred_results/pred_unavailable_files.txt', 'w') as f:
        for item in unavailabel:
            f.write(f"{item}\n")
def main():
    parser = argparse.ArgumentParser()
    parser.add_argument('--path_data', type=str, required=True,
                        help='Folder containing imagery and geojson labels')
    parser.add_argument('--output_mask_path', required=True, type=str,
                        help='Path to save output masks')
    parser.add_argument('--buffer_meters', default=2, type=float,
                        help='Buffer distance (meters) around graph')
    parser.add_argument('--burnValue', default=255, type=int,
                        help='Value of road pixels (for plotting)')
    parser.add_argument('--overwrite_ims', default=1, type=int,
                        help='Switch to overwrite 8bit images and masks')
    parser.add_argument('--is_SN5', action='store_true', help='SN3 or SN5')

    args = parser.parse_args()
    output_mask_path = args.output_mask_path
    if not os.path.isdir(output_mask_path):
        os.makedirs(output_mask_path, exist_ok=True)
    is_SN5 = args.is_SN5
    print('is_SN5->', is_SN5)
    dir_path = args.path_data
    print('doing...', dir_path)
    create_masks(dir_path,
                 buffer_meters=args.buffer_meters,
                 is_SN3=not is_SN5,
                 burnValue=args.burnValue,
                 output_mask_path=output_mask_path,
                 make_plots=0,
                 overwrite_ims=1)
if __name__ == "__main__":
    # Set up the arguments for testing
    sys.argv = [
        'create_road_masks.py',
        '--path_data', 'benchmark/datasets/SpaceNet8/SpaceNet8/baseline/data',
        '--output_mask_path', 'pred_results/pred_road_masks'
    ]
    
    # Run the main function
    main()
\end{lstlisting}
\caption{Road mask generation code solution (continued, page 3 of 3)}
\label{list:full-example-p3}
\end{listing*}

%% file: tables/licenses.tex
\begin{table}[htbp]
\small
\centering
\setlength{\tabcolsep}{4pt}
\begin{tabular}{lc}
\toprule
\textbf{License} & \textbf{Repositories} \\
\midrule
MIT & 449 \\
GNU & 247 \\
Apache & 145 \\
BSD & 84 \\
CC & 57 \\
Boost & 4 \\
Public Domain & 3 \\
ISC & 1 \\
Eclipse & 1 \\
PolyForm & 1 \\
Mulan & 1 \\
Other & 15 \\
\bottomrule
\end{tabular}
\caption{License information for repositories used in \dataset.}
\label{tab:licenses}
\vspace{-11pt}
\end{table}

\begin{table}[htbp]
\small
\centering
\setlength{\tabcolsep}{4pt}
\begin{tabular}{lc}
\toprule
\textbf{Repositories}\\
\midrule
GabrieleLozupone/AXIAL \\
fhalab/MLDE \\
snacktavish/TreeToReads \\
usnistgov/SDNist \\
ruppinlab/CSI-Microbes-identification \\
fenchri/edge-oriented-graph \\
SNU-LIST/QSMnet \\
Ramprasad-Group/polygnn \\
gdalessi/OpenMORe \\
svalkiers/clusTCR \\
AI-sandbox/SALAI-Net \\
pixelite1201/agora\_evaluation \\
jsunn-y/PolymerGasMembraneML \\
spectrochempy/spectrochempy \\
usnistgov/atomgpt \\
\bottomrule
\end{tabular}
\caption{Repositories with other licenses.}
\label{tab:other-licenses}
\vspace{-11pt}
\end{table}

%% file: expert_questionnaire.tex
\begin{table*}[t]
\begin{tcolorbox}[title=\dataset Quality Evaluation]
\small
Hello domain experts and thank you for collaborating with us on \pipeline! We have taken your feedback from the pilot study into account and aimed to improve the quality of the tasks generated through our pipeline. \newline

In this round of validation, we aim to rigorously evaluate the quality of the tasks in the training set through three dimensions: the task instruction, code solution, and task difficulty. \newline

You will be given a folder containing three components: the program (.py file), the task instruction and link to original GitHub file (metadata.jsonl), and zip file containing the program dependencies and output in the folder gold\_results. After examining all these components please answer the following questions. \newline 

\textbf{Expert Name.} [    ]\newline

\textbf{Task ID.} [    ] \newline

\textbf{Task Instruction.}\newline

In this section you will be assessing the quality of the task instruction by determining whether it is meaningful and realistic, correctly expressed in the domain scientific language, and clear.\newline

1. Is this a meaningful and realistic scientific data analysis task that a scientist in your field would perform in their research? \textbf{[Yes/No]} \newline

2. Is the instruction correctly expressed in the domain scientific language? \textbf{[Yes/No]} \newline

3. Is the instruction clear and contains all required information needed to complete the task - goal, methods, input, and output? In other words, if you were given this instruction as a task, would you have the information you need to start writing a solution? \textbf{[Yes/No]} \newline

4. If your answer to the previous question was ``No'', what is missing? \newline

\textbf{Program Solution.}\newline 

In this section you will answer two questions about the functionality equivalence with the original program on GitHub and the program correctness.\newline

1. Does the program perform the same functionality as the original program on GitHub? There might be changes to the program in the adaptation process to make it executable in a standalone environment (e.g. changes to the import statements, to the input / output routines, etc.) Please ignore all these stylistic changes and focus on whether the core functionality of the program remains unchanged. \textbf{[Yes/No]} \newline

2. Does the program represent a valid solution to the task? There could be multiple possible solutions to a task, here you should just determine whether the program is a valid solution and correctly addresses the goal of the task.\textbf{ [Yes/No]} \newline

\textbf{Task Difficulty.} \newline

In this section you will answer rate the task difficulty.\newline

1. How would you rate the difficulty level of the task? Your judgment about the task difficulty should be realistic ( i.e. do not assume familiarity with certain libraries/methods/packages.). In other words, if you had to write a solution for the task right now, how long would that take you? As a general rule of thumb, tasks that can be completed within 15 min are considered easy, those requiring a duration from 15 min to 1 hr are considered medium, and those requiring 1+ hrs are hard. \textbf{[Easy/Medium/Hard]} \\
\end{tcolorbox}
\caption{Questionnaire shared with domain experts for quality evaluation}
\label{box:questionnaire}
\end{table*}